\documentclass[letterpaper, conference]{IEEEtran}
\IEEEoverridecommandlockouts

\usepackage{cite}
\usepackage{amsmath,amssymb,amsfonts}

\usepackage{graphicx}
\usepackage{textcomp}
\usepackage{xcolor}
\usepackage{booktabs}
\usepackage{amsmath,amssymb,mathtools}
\usepackage[ruled,vlined,linesnumbered]{algorithm2e}
\SetKwInput{KwIn}{Input}
\SetKwInput{KwOut}{Output}

\usepackage{float}
\usepackage{bbm}
\usepackage{cuted}
\usepackage{capt-of}
\usepackage{siunitx}
\usepackage{xcolor}
\usepackage{multirow}
 \usepackage{hyperref}

\usepackage{graphicx}
\usepackage{etoolbox}
\usepackage{capt-of}

\def\BibTeX{{\rm B\kern-.05em{\sc i\kern-.025em b}\kern-.08em
    T\kern-.1667em\lower.7ex\hbox{E}\kern-.125emX}}

\def\BibTeX{{\rm B\kern-.05em{\sc i\kern-.025em b}\kern-.08em
    T\kern-.1667em\lower.7ex\hbox{E}\kern-.125emX}}

\newcommand{\vect}[1]{\mathbf{#1}}

\usepackage{mathtools}

\begin{document}

\title{\vspace*{4mm}Kinematify: Open-Vocabulary Synthesis of High-DoF Articulated Objects\\

\thanks{$^{1}$Deemos Corporation, Wilmington, DE 19801, USA. Emails: \texttt{\{joel.wang, dingyou, zhangqx\}@deemos.com}.}%
\thanks{$^{2}$ShanghaiTech University, Shanghai, China. Emails: \texttt{\{wangdy2024, zhangqx1, yujingyi, xulan1\}@shanghaitech.edu.cn}.}%
\thanks{$^{3}$Contextual Robotics Institute, UC San Diego, La Jolla, CA 92093, USA. Emails: \texttt{\{jiw179, jih189\}@ucsd.edu}.}%
}

\author{
Jiawei Wang$^{1,3}$ \quad
Dingyou Wang$^{1,2}$ \quad
Jiaming Hu$^{3}$ \quad
Qixuan Zhang$^{1,2}$\,$^\dagger$ \quad
Jingyi Yu$^{2}$\,$^\ast$ \quad
Lan Xu$^{2}$\,$^\ast$%
}

\newcommand{\insertteaser}{%
  \begin{center}
    \includegraphics[width=\textwidth]{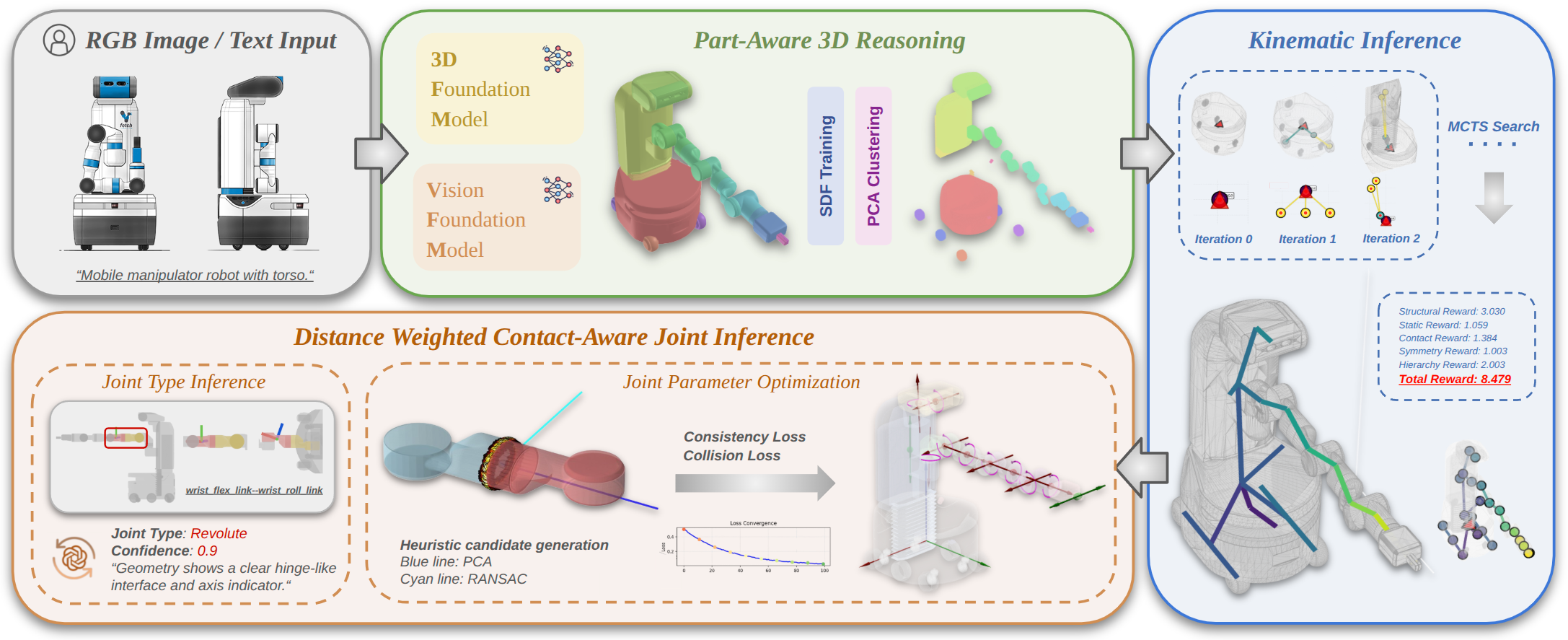}
    \captionof{figure}{\textbf{Overview of Kinematify.} A part-aware 3D foundation model first reconstructs a segmented digital twin. Then, the kinematic tree is recovered via Monte Carlo Tree Search (MCTS) driven by rewards for structure, stability, contact, symmetry, and hierarchy. Finally, joint types are predicted by a vision language model (VLM), and joint parameters are optimized on the parent link’s signed distance field (SDF) to enforce contact consistency and avoid collisions.}
    \label{fig:teaser}
  \end{center}%
}

\IEEEaftertitletext{\vspace{-0.3\baselineskip}\insertteaser}

\maketitle

\begingroup
\renewcommand\thefootnote{\fnsymbol{footnote}}
\footnotetext[2]{Project lead: Qixuan Zhang (\texttt{zhangqx@deemos.com}).}
\footnotetext[1]{Corresponding authors: Jingyi Yu (\texttt{yujingyi@shanghaitech.edu.cn}), 
Lan Xu (\texttt{xulan1@shanghaitech.edu.cn}).}
\endgroup


\begin{abstract}
A deep understanding of kinematic structures is essential for robot motion and interaction with the environment. Such understanding is captured through articulated objects, which are essential for physical simulation, motion planning, and policy learning. However, creating these models, particularly for objects with high degrees of freedom (DoF), remains a significant challenge. Existing methods typically rely on motion sequences or strong assumptions from hand-curated datasets. In this paper, we introduce Kinematify, an automated framework that synthesizes articulated objects from arbitrary RGB images or textual descriptions. Our method addresses two core challenges: (i) inferring kinematic topologies for high-DoF objects and (ii) estimating joint parameters from static 3D geometry. To achieve this, we combine MCTS search for structural inference with geometry-driven optimization for joint reasoning, producing physically consistent and functionally valid models.
We evaluate Kinematify on diverse inputs from both synthetic environments and real-world, demonstrating improvements in registration and kinematic topology accuracy over prior work. 
\url{https://sites.google.com/deemos.com/kinematify}
\end{abstract}


\section{Introduction}
Enabling robots to effectively interact with objects, as well as to model their own articulated structures for self-perception and adaptation, requires an accurate understanding of kinematic topologies and joint parameters. Articulated robot descriptions capture this understanding by encoding geometry, kinematic dependencies, and dynamic constraints in standard formats like the Unified Robot Description Format (URDF) \cite{ros_wiki_urdf_2023}. These descriptions are essential for robotic tasks such as manipulation, locomotion, and policy learning. However, customizing such descriptions for articulated objects remains a significant challenge, demanding substantial manual effort, especially for high degrees of freedom (DoF) systems like humanoids, quadrupeds, and arms. This difficulty arises from the labor intensive processes of part-aware 3D modeling, resolving intricate kinematic dependencies, and inferring precise joint parameters.

While recent advances in part-aware 3D generation \cite{zhang2025bang, yang2025holopart, tang2024partpacker, gao2022get3d} now enable the on-demand creation of high-quality segmented meshes from RGB images or textual descriptions, the bottleneck of kinematics inference remains. This challenge has driven robotics researchers to explore high-DoF articulated objects generation approaches.

Prior work has followed two main directions. Geometry-first approaches infer parts and joints from dense 4D sequences or multi-scan data, which achieve high fidelity but rely on controlled capture settings \cite{liu2023reart,huang2021multibodysync}. Program-synthesis pipelines, in contrast, predict executable descriptions directly from visual inputs \cite{real2code,articulateanything,chen2024urdformer}. While effective, these systems mainly target everyday objects such as laptops, bottles, and drawers, which typically contain only a few moving parts and relatively simple kinematic dependencies. In the context of self-modeling, related work such as AutoURDF \cite{lin2024autourdf} extends this idea to robots, recovering topology and joint types from point-cloud sequences. However, it presumes motion data and is largely limited to serial-chain structures, whereas high-DoF objects often exhibit multi-branched linkages.

To address these challenges, we introduce \textbf{Kinematify}, a framework that generates articulated 3D objects from RGB images or texts. An overview of Kinematify is shown in Fig.~\ref{fig:teaser}. Kinematify generates the segmented mesh with a part-aware 3D foundation model, such as BANG \cite{zhang2025bang}, then infers the kinematic tree using an MCTS \cite{coulom2006mcts, kocsis2006uct} objective that balances hierarchy and structural regularity. Subsequently, it estimates joint parameters via \textbf{DW-CAVL}, a novel \textbf{D}istance-\textbf{W}eighted \textbf{C}ontact-\textbf{A}ware \textbf{V}irtual \textbf{L}inkage optimization approach. This approach preserves near-contact regions while penalizing collisions under virtual motion. The resulting description is exported to URDF and is readily convertible to formats like MJCF \cite{todorov2012mujoco} or USD \cite{openusd_intro_usd_2021}. Our contributions are:

\begin{itemize}
\item \emph{An open-vocabulary articulated object generation framework.} Kinematify generates physics-aware articulated objects directly from arbitrary RGB images or textual descriptions, without requiring motion data, training, or pre-defined articulation priors.
\item \emph{A MCTS-based kinematic tree inference approach.} We propose a search objective that encodes structural priors like hierarchy and regularity to resolve ambiguous attachments for complex, high-DoF articulated objects with multiple branches.
\item \emph{A SDF-driven joint parameter estimation approach.} The DW-CAVL algorithm accurately infers revolute and prismatic joint parameters from static geometry by optimizing an SDF-based, contact-aware objective under virtual motions.
\end{itemize}


\section{Related Work}

\subsection{3D Articulated Object Modelling}
A significant body of work focuses on reconstructing the kinematic structure of everyday objects from visual data.
The most common paradigm leverages motion to reveal articulation \cite{huang2021multibodysync, liu2023reart, liu2025artgsbuildinginteractablereplicas, shen2025gaussianartunifiedmodelinggeometry, lin2024autourdf}. By observing an object over time, these methods can directly infer which parts move together and identify the axes of rotation or translation. For example, MultiBodySync \cite{huang2021multibodysync} registers multiple 3D scans of an object in different states, using spectral synchronization to jointly solve for part segmentation and motion. Similarly, ReArt \cite{liu2023reart} fits a rearticulable model to a 4D point cloud sequence by jointly optimizing segmentation, topology, and joint parameters. These methods achieve high fidelity but depend critically on the availability of multi-view or temporal data, which requires a controlled capture setup.
Another trend frames articulation modelling as a program synthesis problem \cite{chen2024urdformer, real2code, articulateanything, luo2025physpartphysicallyplausiblecompletion, qiu2025articulateanymeshopenvocabulary3d, wu2024genpose, chattopadhyay2024instructpart, Li2024articulateddiffusion, An2024gad, Wang2024genad, Li2023interactron, Abbatematteo2024learning, hong2024akb, hassan2023part-e, yariv2023articulated, wang2023whatisparse}. URDFormer \cite{chen2024urdformer} trains a transformer to predict a URDF from a single image, relying on a large-scale synthetic dataset of image-URDF pairs. Real2Code \cite{real2code} and Articulate-Anything \cite{articulateanything} leverage large language models to generate code-based representations of articulated objects, with the latter using a reinforcement learning loop to refine the model through simulation feedback. While powerful in their open-vocabulary capabilities, these methods often struggle with the multi-branch kinematics, which are common in high-DoF objects.

\subsection{Robot Self-Modeling}
Distinct from general everyday object modeling, robot self-modeling is the online process by which a robot autonomously discovers its own body plan \cite{kwiatkowski2019task, chen2022fully, ledezma2023machine, lin2024autourdf, Fu2024ditto}. This is typically achieved by correlating motor actions with sensory feedback.
The foundational concept of task-agnostic self-modeling \cite{kwiatkowski2019task} involves a robot performing random motions and building a self-representation from the resulting data. This has been realized in various ways. Ledezma et al. \cite{ledezma2023machine} used IMU sensors on each link, applying machine learning to the sensor data to explicitly solve for the robot's topology and kinematic parameters. These methods are powerful but require an embodied agent with access to its own motor and sensory signals. AutoURDF \cite{lin2024autourdf} represents a purely visual approach to this problem. It operates on time-series point cloud frames of a robot in motion, but without access to the underlying motor commands. By tracking the 6-DoF transformations of point clusters, it segments moving parts, infers the kinematic tree using a minimum spanning tree, and estimates joint parameters. It demonstrated superior performance in topology inference for serial-chain objects.

\section{Kinematify}
\label{sec:method}

We introduce Kinematify, a framework that generates kinematics-aware articulated objects directly from RGB images or textual descriptions in a zero-shot context. 

\subsection{Preliminaries}
\label{subsec:prelim}
\textbf{Assemblies and parts.}
An \emph{assembly} $\mathcal{A}$ is a set of parts $P=\{P_i\}_{i=1}^N$. Part $P_i$ has a triangulated surface mesh $M_i=(V_i,F_i)$ with vertices $V_i\subset\mathbb{R}^3$ and triangular faces $F_i\subset\{1,\dots,|V_i|\}^3$. Each part stores a world transform $T_i\in\mathrm{SE}(3)$, an intrinsic rotation $R_i\in\mathrm{SO}(3)$ used for alignment, a centroid $c_i\in\mathbb{R}^3$, a robust volume $v_i>0$, and axis-aligned bounding-box extents $e_i\in\mathbb{R}^3_{>0}$.

\textbf{Graphs and kinematic tree.}
We build an undirected connection graph $G=(V,E)$ with $V\leftrightarrow P$, where an edge indicates geometric contact. A directed kinematic tree $T=(V,E_T)$, rooted at the base link $b\in V$, orients $G$ and annotates joints.

\textbf{Joints.}
For a directed edge $(u\!\to\!v)\in E_T$, a joint stores a type $J_{uv}\in\{fixed,revolute,prismatic\}$, a parent-to-child origin $o_{uv}\in\mathbb{R}^3$, and if movable, an axis $\vect{a}_{uv}\in\mathbb{S}^2$ and optionally a pivot $\vect{p}_{uv}\in\mathbb{R}^3$ when revolute.

\subsection{Part-Aware 3D Representations}
\label{subsec:repr}

We generate part-level 3D meshes with a part-aware 3D foundation model \cite{zhang2025bang} from the input RGB images or textual description, and discard meshes with too few vertices or a degenerate spatial spread.
For each prospective parent part, we train a continuous SDF \cite{park2019deepsdf, gropp2020igr, sitzmann2020siren} $f_\theta:\mathbb{R}^3\to\mathbb{R}$ on (i) noisy surface points, (ii) near-surface offsets, and (iii) far samples in a bounding AABB. 

Afterward, we build a connection graph $G$ with the trained SDF, as shown in the middle left panel of Fig. \ref{fig:pipeline_demo}. Given two candidate parts $A$ and $B$ with respective SDFs $f_{\theta_A}$ and $f_{\theta_B}$, we evaluate mutual distances between their sampled surfaces under $f_{\theta}$. Pairs of parts whose minimum bidirectional distance falls below a tolerance $\epsilon$ are declared in contact, and an undirected edge is added between them.

\subsection{Kinematic Topology Inference}
\label{subsec:topology}

We orient the graph \(G\) into a directed kinematic tree \(T\) with root \(b\).
For any directed tree \(X\), let \(V(X)\) and \(E(X)\) denote its node and oriented edge sets.
Node positions \(c_i\in\mathbb{R}^3\) are reference points of part \(i\).
For an oriented edge \((u\!\to\!v)\in E(X)\), its edge-origin vector is \(o_{uv}:=c_v-c_u\).
For node-wise functions \(f:V(\tilde T)\!\to\!\mathbb{R}\), the average is \(\overline{f}:=|V(\tilde T)|^{-1}\sum_{i\in V(\tilde T)} f(i)\).
Depth \(d(i)\) is the graph distance from the root \(b\) to node \(i\).
The out-degree in the directed tree is \(\deg^+(i)\).
All edge-wise sums \(\sum_{(u\to v)}\) are over \((u\to v)\in E(\tilde T)\) unless stated otherwise.
A positive distance threshold \(d_{\max}>0\) is used when attaching disconnected components.

\paragraph{Base selection and BFS orientation}
We choose the base link \(b\) as any node in \(G\) with the highest undirected degree. Starting from \(b\), we run BFS on \(G\) as a warm start for the MCTS search.
During BFS, when a new neighbour \(v\) is first reached from an already visited node \(u\), we define \(u\) as the parent and \(v\) as the child, orient the edge as \(u\!\to\!v\), and set its origin \(o_{uv}\gets c_v-c_u\).
Any edge whose addition would create a cycle is not inserted into \(T\) and is recorded as broken.
If \(G\) is disconnected, each remaining component is attached to the current tree by adding a virtual edge from its nearest neighbor, provided the Euclidean distance is at most \(d_{\max}\).

\subsubsection{State, actions, constraints}
A search state is \(S=(T_S,V_S,B_S)\), where \(T_S\) is the current partial directed tree, \(V_S\subseteq V(G)\) the visited-node set, and \(B_S\subseteq E(G)\) the set of broken undirected edges discovered so far. An action adds a feasible oriented edge \(u\!\to\!v\) with \(u\in V_S\) and \(v\notin V_S\).
To respect discovered symmetries, we form part clusters \(\{C_k\}\) by the Chamfer distance between segmented meshes. During expansion, we forbid connecting two nodes that belong to the same multi-member cluster to avoid spurious intra-cluster links.

\begin{figure}[t]
\centering
\begin{tabular}{cc}
\includegraphics[width=0.48\textwidth]{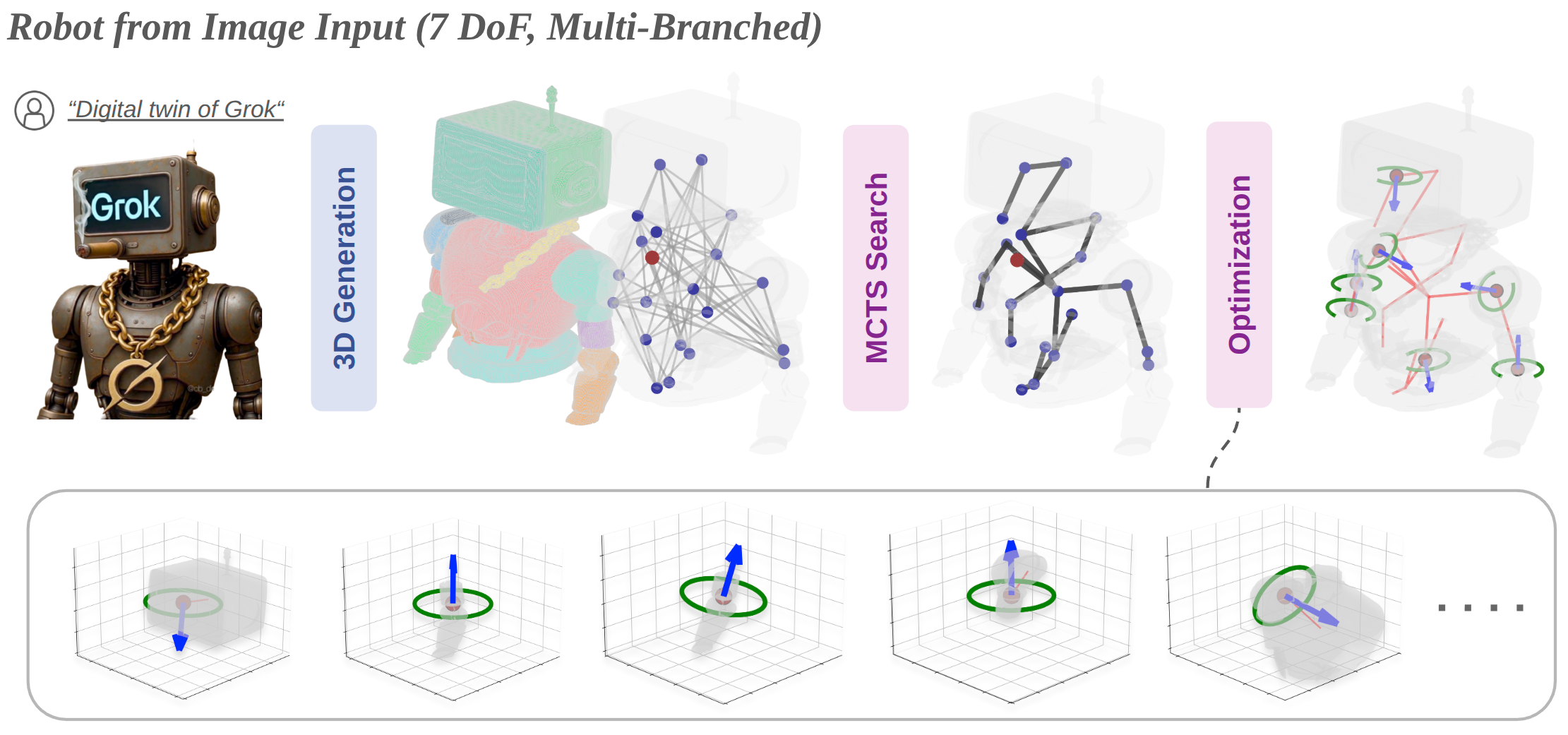} &
\end{tabular}
\caption{Pipeline of Kinematify for recovering articulated robots from a single RGB image. \textbf{Step 1}: A 3D foundation model generates a segmented mesh of the robot. \textbf{Step 2}: A contact graph is constructed over mesh parts, capturing candidate relations between components. \textbf{Step 3}: Infer the kinematic tree using MCTS, resolving ambiguous connections by leveraging structural priors such as hierarchy and symmetry. \textbf{Step 4}: Refine joint parameters using the DW-CAVL optimization approach while preserving near-contact geometry. \textbf{Bottom row}: Examples of inferred revolute joints with optimized axes.}
\label{fig:pipeline_demo}
\end{figure}

\subsubsection{Transition}
Applying an action \(u\!\to\!v\) updates the edge origin \(o_{uv}\gets c_v-c_u\), provisionally treats the joint as fixed until later typing, inserts \(v\) into \(V_S\), and appends to \(B_S\) any undirected edge \((v,w)\in E(G)\) with \(w\in V_S\setminus\{u\}\) that would otherwise form a cycle.

\subsubsection{Reward}
For a completed tree \(\tilde T\), the terminal reward is a weighted sum of five terms:

\paragraph{\(R_{\mathrm{struct}}\)}
This term penalizes large depth variance and high degree deviation:
\begin{equation}
R_{\mathrm{struct}}
=\frac{1}{\overline{d^2}+\overline{(\deg^+\!-\!k)^2}},
\end{equation}
where \(k\) is the preferred out-degree and \(\lambda>0\) is a hyperparameter.

\paragraph{\(R_{\mathrm{static}}\)}
This term favours centre-of-mass support to reduce gravitational torque about joint frames.
Let \(v_i>0\) denote the estimated volume of part \(i\) and \(m_i=\rho\,v_i\) its mass for a density parameter \(\rho>0\).
With subtree mass \(M(i)\) and subtree centre \(\mathbf{c}_{\mathrm{sub}}(i)\),
\begin{align}
M(i)&=m_i+\sum_{j\in \mathrm{ch}(i)} M(j),\\
\mathbf{c}_{\mathrm{sub}}(i)&=\frac{m_i\,c_i+\sum_{j\in \mathrm{ch}(i)} M(j)\,\mathbf{c}_{\mathrm{sub}}(j)}{M(i)},
\end{align}
where \(\mathrm{ch}(i)\) is the set of children of \(i\) in \(\tilde T\).
Let \(\hat{\mathbf z}^{-}=[0,\,0,\,-1]^\top\) denote downward unit gravity and \(g>0\) the gravitational constant.
The total gravitational torque is
\begin{align}
\tau &= \sum_{i\ne b}\left\| \big(\mathbf{c}_{\mathrm{sub}}(i)-c_i\big)\times \big(M(i)\,g\,\hat{\mathbf z}^{-}\big)\right\|_2,\\
R_{\mathrm{static}} &= \frac{1}{1+\tau/\sigma_\tau},
\end{align}
with \(\sigma_\tau>0\) a robust per-assembly normaliser based on MAD scale.

\paragraph{\(R_{\mathrm{contact}}\)}
We quantify contact strength from SDF-based bidirectional proximity. Let \(s(u,v)\in[0,1]\) denote the contact strength of a physical contact on edge \((u\!\to\!v)\).
We reward higher average strength:
\begin{equation}
R_{\mathrm{contact}}=\frac{1}{|E(\tilde T)|}\sum_{(u\to v)\in E(\tilde T)} s(u,v).
\end{equation}

\paragraph{\(R_{\mathrm{sym}}\)}
Within each discovered symmetry cluster \(C_k\) (\(|C_k|\ge 2\)), we prefer equal depths and a shared parent, such as legs attached to the same torso, fingers to the same palm.
Let \(P_k=\{\mathrm{parent}(i): i\in C_k,\, i\ne b\}\).
We define
\begin{equation}
\begin{aligned}
S_k &= \frac{1}{1+\mathrm{Var}\!\big(d(i): i\in C_k\big)}
      \;+\;\Big[1 - \frac{|P_k|-1}{|C_k|-1+\varepsilon}\Big],\\
R_{\mathrm{sym}} &= \mathrm{mean}_k\, S_k .
\end{aligned}
\end{equation}
The second term equals \(1\) when all parts in \(C_k\) share the same parent (\(|P_k|=1\)) and decreases linearly as parents diversify. \(\varepsilon>0\) avoids division by zero.

\paragraph{\(R_{\mathrm{hier}}\)}
We discourage children much larger than their parents by estimated volume.
With a small \(\varepsilon>0\) to avoid divide-by-zero,
\begin{equation}
R_{\mathrm{hier}}=\frac{1}{1+\sum_{(u\to v)} \max\!\left\{0,\ \frac{v_v}{v_u+\varepsilon}-1\right\}}.
\end{equation}

\subsubsection{Search}
We use Monte Carlo Tree Search (MCTS) with UCT. Each state is \(S=(T_S,V_S,B_S)\). From a state \(S\), each child \(c\) corresponds to applying one feasible action \(u\!\to\!v\).
Let \(Q(c)\) be the cumulative return backed up through child \(c\), \(N(c)\) its visit count, and \(N(\mathrm{parent})\) the visit count of its parent state.
With exploration constant \(C>0\), selection chooses
\begin{equation}
\arg\max_{c}\ \frac{Q(c)}{N(c)}\;+\; C\sqrt{\frac{\ln N(\mathrm{parent})}{N(c)}}.
\end{equation}
Rollouts greedily complete the tree by repeatedly choosing any available action with the highest immediate score, and the terminal return \(R(\tilde T)\) is backed up along the simulation path.
This objective helps resolve symmetric attachments and multi-branch ambiguities at scale. The middle right panel of Fig. \ref{fig:pipeline_demo} shows the kinematics structure after MCTS search.

\begin{algorithm}[t]
\caption{\textsc{Kinematify}}
\KwIn{Connection graph $G=(V,E)$, base $b$, SDFs $\{f_{\theta}\}$, samples $\{P_v\}$}
\KwOut{Kinematic tree $T=(V,E_T)$, joints $\{J_{uv}\}$}

$S_0\!\leftarrow\!(\emptyset,\{b\},\emptyset)$; init stats $Q,N\!\leftarrow\!0$ \;
\For{$t=1$ \KwTo $N_{\max}$}{
  $S\leftarrow S_0$; path $\mathcal{P}\leftarrow\emptyset$ \;
  \While{$|V_S|<|V|$}{
    \If{untried edge exists}{$a\leftarrow$ pick untried;\ $S\leftarrow$ Transition$(S,a)$;\ \textbf{break}}
    \Else{$S\leftarrow \arg\max_{c}\frac{Q(c)}{N(c)}+C\sqrt{\ln N(S)/N(c)}$}
    $\mathcal{P}.\text{append}(S)$
  }
  $\tilde T\leftarrow$ greedy rollout from $S$ \;
  $R\leftarrow$ Reward$(\tilde T)$; backprop $R$ along $\mathcal{P}$ \;
}
$T\leftarrow$ best cached tree \;

\For{edge $(u\!\to\!v)\in E_T$}{
  candidates $\leftarrow$ generate from contact stats \;
  \For{candidate $(p,u)$}{optimize $\mathcal{J}_{rev}$ or $\mathcal{J}_{pri}$}
  select best class: revolute if $s_{rev}>\zeta s_{pri}$, else prismatic\;
  store $J_{uv}$
}
\KwRet{$(T,\{J_{uv}\})$}
\end{algorithm}
\subsection{Joint Reasoning}
\label{subsec:joint-repr}

We render orthographic viewsets for the whole assembly and for joint close-ups. For each $(u\!\to\!v)$, we query VLM on the joint viewset and adopt a decision with abstention. If the VLM successfully identifies the joint type, we group joints by child clusters $C_k$ and select a representative by majority and correct outliers.

Let $P_A=\{\mathbf{a}_i\}$ and $P_B=\{\mathbf{b}_j\}$ be surface samples of two parts $A$ and $B$ in a common frame. We first extract a contact region $\mathcal{C}$ as the union of points on either part whose nearest neighbour on the other part lies within a small threshold. To downweight spurious pairs, each $\mathbf{x}\in\mathcal{C}$ is assigned a weight that decays with its nearest-neighbour distance, using these weights we compute a weighted contact centroid $\boldsymbol{\mu}_c$ and a weighted covariance $\boldsymbol{\Sigma}$. The principal direction with the smallest variance provides a hinge-axis estimate $\widehat{\mathbf{u}}_{\mathrm{PCA}}$. In parallel, we obtain a contact normal $\widehat{\mathbf{n}}$ by averaging nearest-point difference vectors across the two surfaces.

We then form a diverse set of revolute candidates $(\mathbf{p},\mathbf{u})$ as follows. Pivots are initialised at the contact centroid ($\mathbf{p}=\boldsymbol{\mu}_c$). Axes are drawn from a compact pool that includes $\widehat{\mathbf{u}}_{\mathrm{PCA}}$, the contact normal $\widehat{\mathbf{n}}$, an orthogonal completion $\widehat{\mathbf{u}}_{\perp}$, the principal axes of $\boldsymbol{\Sigma}$, and a few random unit directions. Along each candidate axis $\mathbf{u}$ we place a handful of pivot samples by sliding $\mathbf{p}$ slightly along $\mathbf{u}$ around $\boldsymbol{\mu}_c$.

\begin{figure*}[!t]
\centering
\includegraphics[width=0.95\textwidth]{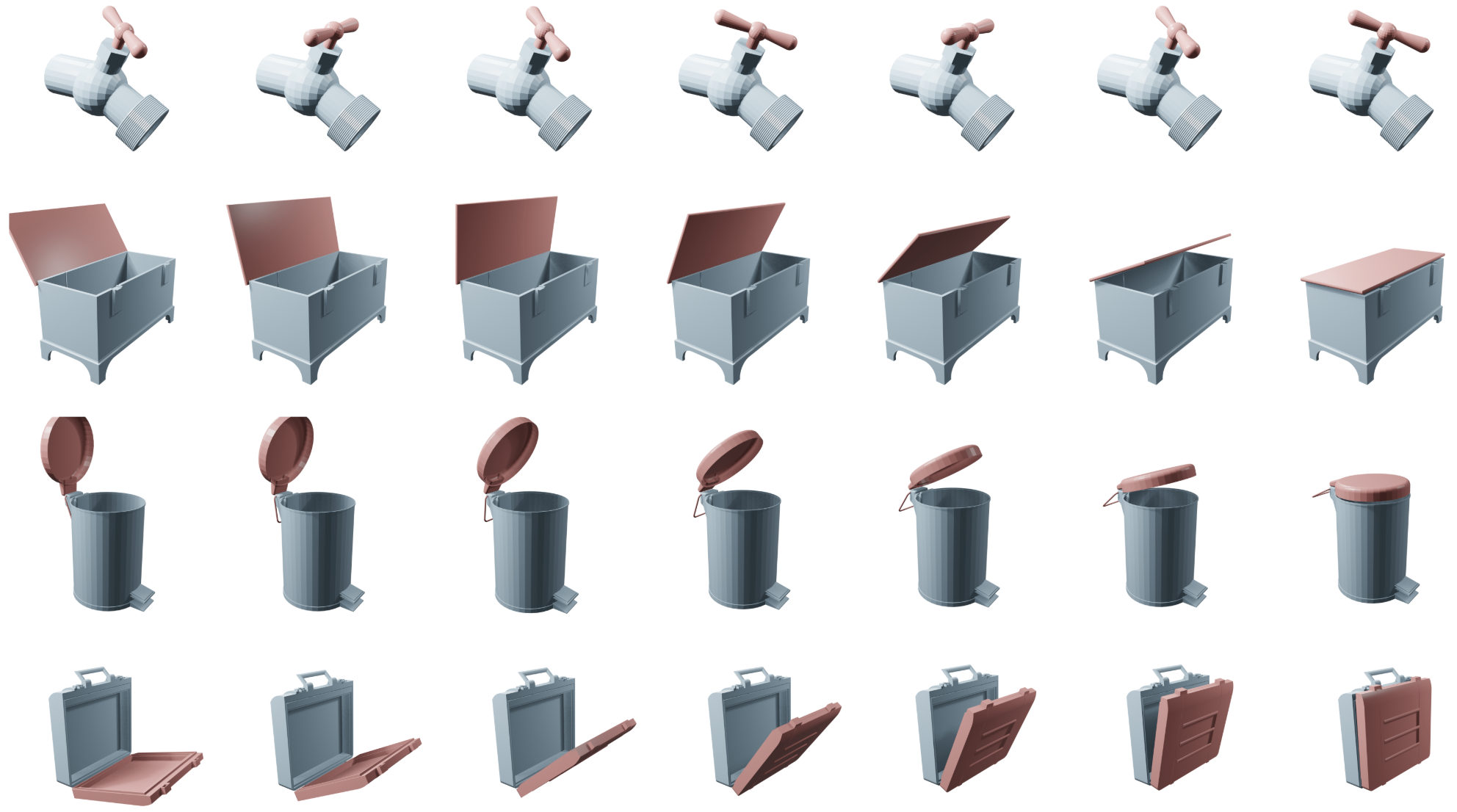}
\caption{\textbf{Examples of articulated objects generated by Kinematify.} Each row shows different objects across a sequence of joint configurations.}
\label{fig:eo_demo}
\end{figure*}

\subsubsection{Differentiable rigid motions}
For a revolute motion $(\mathbf{p},\mathbf{u},\theta)$ with $\|\mathbf{u}\|_2=1$,
\begin{align}
\mathbf{y} &= \mathbf{p}+\mathbf{R}(\mathbf{u},\theta)\,(\mathbf{x}-\mathbf{p}), \\
\mathbf{R}(\mathbf{u},\theta) &= \cos\theta\,\mathbf{I}+\sin\theta\,[\mathbf{u}]_\times+(1-\cos\theta)\,\mathbf{u}\mathbf{u}^\top,
\end{align}
where $[\mathbf{u}]_\times$ is the $3\times3$ cross-product matrix.
For a prismatic motion with displacement $t\in\mathbb{R}$, $\mathbf{y}=\mathbf{x}+t\,\mathbf{u}$.
We parametrise $\mathbf{u}$ by an unconstrained $\mathbf{a}\in\mathbb{R}^3$ via $\mathbf{u}=\mathbf{a}/\|\mathbf{a}\|_2$, whose Jacobian is
\begin{equation}
\frac{\partial \mathbf{u}}{\partial \mathbf{a}}=\frac{1}{\|\mathbf{a}\|_2}\left(\mathbf{I}-\frac{\mathbf{a}\mathbf{a}^\top}{\|\mathbf{a}\|_2^2}\right).
\end{equation}

\subsubsection{DW-CAVL objective}
Let child samples be $\{\mathbf{b}_i\}$.
A virtual motion parameter $\delta\in\Theta$ denotes either a revolute angle $\theta$ or a prismatic displacement $t$.
Let $\Phi_\delta$ be the corresponding rigid transform of the child, and define
\[
s_0(\mathbf{x})=f_\theta(\mathbf{x}),\qquad
s_\delta(\mathbf{x})=f_\theta\!\big(\Phi_\delta(\mathbf{x})\big).
\]
With hyperparameters volumetric margin $m_{\mathrm{vol}}>0$, logistic sharpness $k>0$, contact band width $\sigma_c>0$, and $\varepsilon_{\mathrm{small}}>0$,
\begin{align}
w_{\mathrm{vol}}(s_0)&=\sigma\!\big(-k(s_0-m_{\mathrm{vol}})\big),\quad \sigma(z)=\frac{1}{1+e^{-z}},\\
w_{\mathrm{dist}}(s_0)&=\exp\!\left(-\frac{s_0^2}{2\sigma_c^2}\right),\qquad w_i=w_{\mathrm{vol}}(s_0(\mathbf{b}_i))\,w_{\mathrm{dist}}(s_0(\mathbf{b}_i)).
\end{align}
The consistency term penalizes separation near contact after motion,
\begin{equation}
\mathcal{L}_{\mathrm{cons}}(\delta)=\frac{\sum_i w_i \big[\max\{0,\ s_\delta(\mathbf{b}_i)-m_{\mathrm{vol}}\}\big]^2}{\sum_i w_i+\varepsilon_{\mathrm{small}}}.
\label{eq:cons}
\end{equation}
Using inverse volumetric weights $\tilde w_i=\sigma\!\big(+k(s_0(\mathbf{b}_i)-m_{\mathrm{vol}})\big)$, the collision term penalises penetration,
\begin{equation}
\mathcal{L}_{\mathrm{coll}}(\delta)=\frac{\sum_i \tilde w_i \big[\max\{0,\ -s_\delta(\mathbf{b}_i)-m_{\mathrm{vol}}\}\big]^2}{\sum_i \tilde w_i+\varepsilon_{\mathrm{small}}}.
\label{eq:coll}
\end{equation}
For revolute joints we regularise the pivot toward $\boldsymbol{\mu}_c$:
\begin{equation}
\mathcal{L}_{\mathrm{reg}}=\lambda_p \|\mathbf{p}-\boldsymbol{\mu}_c\|_2^2,\qquad \lambda_p\ge 0.
\end{equation}
Aggregating over $\delta\in\Theta$ yields
\begin{equation}
\mathcal{J}(\mathbf{p},\mathbf{u})=\frac{1}{|\Theta|}\sum_{\delta\in\Theta}\big(\lambda_c \mathcal{L}_{\mathrm{cons}}(\delta)+\lambda_{\mathrm{coll}} \mathcal{L}_{\mathrm{coll}}(\delta)\big)+\mathcal{L}_{\mathrm{reg}},
\label{eq:total-loss}
\end{equation}
with nonnegative weights $\lambda_c,\lambda_{\mathrm{coll}}$.

\subsubsection{Candidate selection}
We rank many candidates $(\mathbf{p},\mathbf{u})$ on subsampled points using $s(\mathbf{p},\mathbf{u})$, refine the top-K on full points by minimising $\mathcal{J}$, and output scores $1/(1+\mathcal{J})$.

In summary, we infer joint parameters from static geometry and select the candidate with the highest score. The bottom panel in Fig. \ref{fig:pipeline_demo} shows the optimized results for each joint for the input.

\begin{table*}[t!]
\centering
\setlength{\tabcolsep}{6pt}
\caption{Unified baseline comparison across robots sorted by degrees of freedom (DoF). 
We report Axis Angle Error (°), Axis Position Error (m), and Tree Edit Distance, where lower values are better.}
\begin{tabular}{l l|c c c c c c c c}
\toprule
\multirow{2}{*}{Metric} & \multirow{2}{*}{Method} &
\multicolumn{1}{c}{Everyday Object} &
\multicolumn{1}{c}{UR10e} &
\multicolumn{1}{c}{Franka Panda} &
\multicolumn{1}{c}{Unitree Go2} &
\multicolumn{1}{c}{Fetch} &
\multicolumn{1}{c}{Allegro} &
\multicolumn{1}{c}{Unitree H1} &
\multirow{2}{*}{Mean}\\
& & {\scriptsize 1-8 DoF} & {\scriptsize 6 DoF} & {\scriptsize 7 DoF} & {\scriptsize 12 DoF} & {\scriptsize 13 DoF} & {\scriptsize 16 DoF} & {\scriptsize 19 DoF} & \\
\midrule
\multirow{3}{*}{$\mathrm{Axis\ Angle\ Error}\!\downarrow$}
& Articulate Anymesh & 35.80 & 39.67 & 42.10 & 53.23 & 75.60 & 78.77 & 79.35 & 57.79 \\
& ArtGS              & \underline{13.80} & \underline{25.52} & \underline{21.30} & \underline{22.32} & \underline{53.81} & \underline{65.59} & \underline{41.29} & \underline{34.80} \\
& Ours               & \textbf{2.92} & \textbf{5.34} & \textbf{10.42} & \textbf{9.97} & \textbf{23.10} & \textbf{31.39} & \textbf{29.31} & \textbf{16.06} \\
\midrule
\multirow{3}{*}{$\mathrm{Axis\ Pos\ Error}\!\downarrow$}
& Articulate Anymesh & \textbf{0.19} & \textbf{0.25} & \underline{0.31} & \underline{0.41} & \underline{0.89} & \underline{0.32} & \underline{0.74} & \underline{0.44} \\
& ArtGS              & 0.75 & 0.97 & 0.68 & 1.13 & 1.93 & 0.67 & 1.32 & 1.06 \\
& Ours               & \underline{0.23} & \underline{0.27} & \textbf{0.15} & \textbf{0.30} & \textbf{0.71} & \textbf{0.21} & \textbf{0.68} & \textbf{0.36} \\
\midrule
\multirow{2}{*}{$\mathrm{TED}\!\downarrow$}
& AutoURDF           & \underline{0.27} & \textbf{0.87} & \underline{1.28} & \underline{2.21} & \underline{3.21} & \underline{4.83} & \underline{8.13} & \underline{2.97} \\
& Ours               & \textbf{0.13} & \underline{1.03} & \textbf{0.89} & \textbf{1.97} & \textbf{1.78} & \textbf{1.22} & \textbf{2.23} & \textbf{1.32} \\
\bottomrule
\end{tabular}
\label{tab:unified}
\end{table*}

\begin{figure}[t]
\centering
\includegraphics[width=0.48\textwidth]{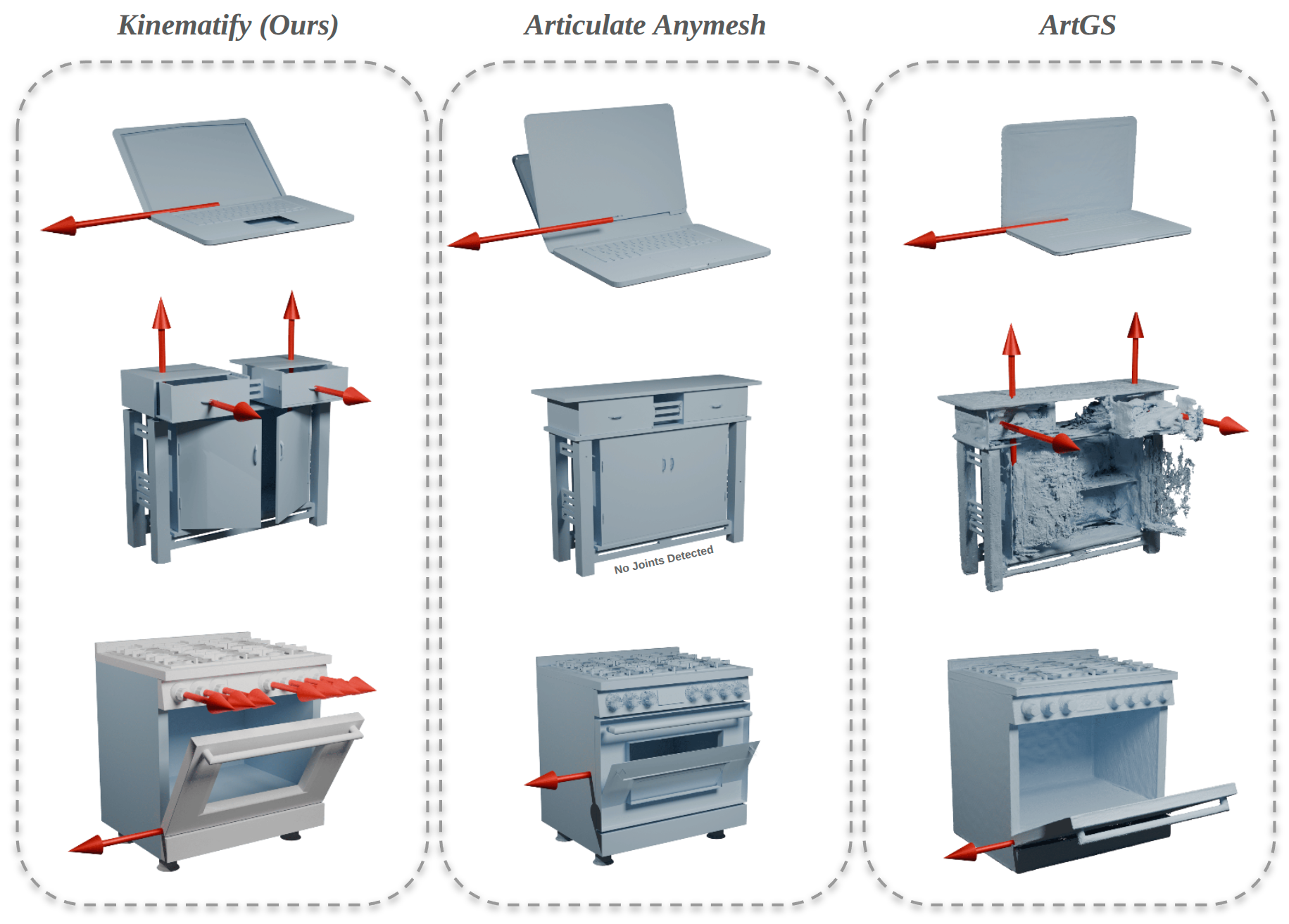}
\caption{\textbf{Qualitative comparison of articulation recovery on everyday objects across three methods:} Kinematify (ours), Articulate Anymesh, and ArtGS. The red line indicates the joint direction.}
\label{fig:comparison_eo}
\end{figure}

\section{Experiments}
\label{sec:exp}

We evaluate Kinematify in two settings: (i) everyday articulated objects and (ii) robotic platforms. We follow prior protocols of Articulate Anymesh~\cite{qiu2025articulateanymeshopenvocabulary3d}, which use ground-truth segmented meshes from PartNet-Mobility \cite{Xiang_2020_SAPIEN, mo2019partnet} to isolate the impact of 3D segmentation. This ensures fair, direct comparisons to baselines that do not accept raw images or texts. We erase the provided kinematic graphs and joint parameters, and reconstruct the mesh. Under this configuration, we compare against the baselines Articulate AnyMesh~\cite{qiu2025articulateanymeshopenvocabulary3d} and ArtGS~\cite{liu2025artgsbuildinginteractablereplicas}.

For the robotics platforms, we evaluate Kinematify on six commonly used robot models spanning a range of DoF. Because ArtGS and Articulate AnyMesh do not expose explicit kinematic tree structure, we additionally compare against AutoURDF~\cite{lin2024autourdf} for kinematics reconstruction performance.

We also conduct experiments on the end-to-end pipelines, starting from RGB images, and evaluate the output quality with ground-truth to quantify the discrepancies.

\subsection{Metrics}
\label{sec:metrics}

We report three metrics evaluating both joint parameter and kinematics tree quality.

\textbf{Axis Angle Error}: The angular deviation between the predicted and ground-truth joint-axis directions, and opposite directions are treated as equivalent.

\textbf{Axis Position Error}: The Euclidean distance between predicted and ground-truth pivot positions in the dataset coordinate frame.

\textbf{Tree Edit Distance}: The Tree Edit Distance~\cite{10.1145/2699485} between the predicted and ground-truth kinematic trees, for instance, the minimal number of node insertions, deletions, or relabelings needed to match the trees.

\begin{figure}[ht]
\centering
\begin{tabular}{cc}
\includegraphics[width=0.47\textwidth]{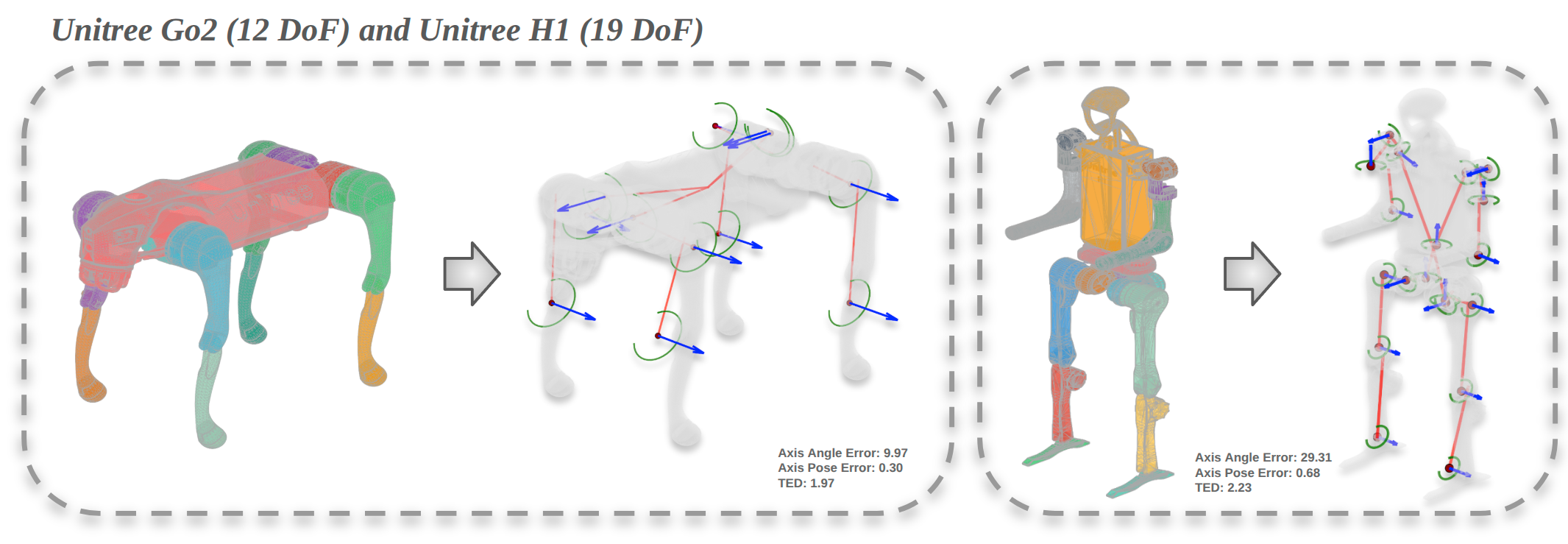} &
\end{tabular}
\caption{\textbf{Demonstration of Kinematify on two high-DoF robots:} Unitree Go2 (12 DoF, left) and Unitree H1 (19 DoF, right). For each case, the pipeline starts from a segmented mesh, followed by kinematic tree inference and joint parameter optimization.}
\label{fig:robot_demo}
\end{figure}

\subsection{Quantitative results}
\textbf{Everyday Objects.}
Table~\ref{tab:unified} reports a comparison of Kinematify against Articulate Anymesh and ArtGS on the PartNet-Mobility benchmark. Our method achieves the lowest axis angle error among all approaches, indicating superior accuracy in joint orientation estimation. In terms of axis position error, Kinematify also performs competitively, with values close to the best baseline. A qualitative visualization of these comparisons is provided in Fig.~\ref{fig:comparison_eo}. Together, these results demonstrate that Kinematify produces precise joint axes and stable pivot placements for everyday objects.

\textbf{Robots.}
We further evaluate performance on six robotic systems by measuring both registration quality and body topology reconstruction. As shown in Table~\ref{tab:unified}, Kinematify reduces the Tree Edit Distance by a substantial margin on average, reflecting more faithful recovery of kinematic structures. Representative results on Unitree H1 and Go2 are visualized in Fig.~\ref{fig:robot_demo}. These findings highlight the effectiveness of our MCTS-based objective in reasoning about high-DoF, multi-branched kinematic structures, surpassing prior methods in structural consistency.

\subsection{End-to-end evaluation}
\label{sec:e2e}

We further evaluate Kinematify in a full end-to-end setting that starts from single RGB images. A part-aware 3D foundation model~\cite{zhang2025bang} first produces a segmented mesh. Then, we apply the kinematic reasoning stack unchanged. Because existing baselines do not natively support image to articulated objects at comparable scope, we report absolute performance rather than head to head comparisons.

\begin{table}[t]
\centering
\setlength{\tabcolsep}{6pt}
\caption{End-to-end results. Numbers are absolute.}
\label{tab:e2e}
\begin{tabular}{l|c c c}
\toprule
Metric & Everyday Objects & Fetch & Panda \\
\midrule
$\mathrm{Axis\ Angle\ Error}\!\downarrow$ & 3.78 & 32.84 & 14.08 \\
$\mathrm{Axis\ Position\ Error}\ (\mathrm{m})\!\downarrow$ & 0.28 & 0.95 & 0.22 \\
$\mathrm{TED}\!\downarrow$ & 0.67 & 2.95 & 1.17 \\
\bottomrule
\end{tabular}
\end{table}

As shown in Table \ref{tab:e2e}, compared to the geometry-only track in Table~\ref{tab:unified}, end-to-end errors increase modestly on EO and more noticeably on Fetch and Panda, consistent with their tighter kinematic tolerances.

\subsection{Ablation study}

We quantify the contribution of each core component by comparing the full method against two ablations: (i) removing the DW-CAVL anchor term so that optimization considers only collision avoidance; and (ii) replacing the MCTS-based kinematic inference with a BFS strategy.

\begin{table}[t]
\centering
\setlength{\tabcolsep}{6pt}
\caption{Ablation study on the proposed method, where EO denotes Everyday Object.}
\begin{tabular}{l l|c c c}
\toprule
Metric & Variant & EO & Fetch & Panda \\
\midrule
\multirow{3}{*}{$\mathrm{Axis\ Angle\ Error}\!\downarrow$}
& w/o MCTS        & \underline{4.32} & \underline{28.30} & \underline{10.92} \\
& w/o DW\mbox{-}CAVL & 13.94 & 42.30 & 29.39 \\
& Ours            & \textbf{2.92} & \textbf{23.10} & \textbf{10.42} \\
\midrule

\multirow{3}{*}{$\mathrm{Axis\ Pos\ Error}\!\downarrow$}
& w/o MCTS        & \underline{0.59} & \underline{0.97} & \underline{0.30} \\
& w/o DW\mbox{-}CAVL & 1.34 & 1.82 & 0.97 \\
& Ours            & \textbf{0.23} & \textbf{0.71} & \textbf{0.15} \\
\midrule
\multirow{3}{*}{$\mathrm{TED}\!\downarrow$}
& w/o MCTS        & 0.39 & 3.32 & 2.97 \\
& w/o DW\mbox{-}CAVL & \underline{0.14} & \underline{1.93} & \underline{0.98} \\
& Ours            & \textbf{0.13} & \textbf{1.78} & \textbf{0.89} \\
\bottomrule
\end{tabular}
\label{tab:ablation_masked}
\end{table}

As shown in Table~\ref{tab:ablation_masked}, substituting MCTS with BFS consistently yields larger TED across robots. BFS greedily attaches along local contacts and lacks long range regularization, leading to incorrect parent choices in symmetric substructures and unbalanced trees. In contrast, removing the DW\mbox{-}CAVL anchor does not drastically change the tree but significantly degrades joint parameters. Without an attraction to the contact centroid and near-surface band, the optimizer favors axes that quickly reduce interpenetration yet drift from true pivots. Overall, the full model achieves the best balance.

\begin{figure}[h]
\centering
\includegraphics[width=0.48\textwidth]{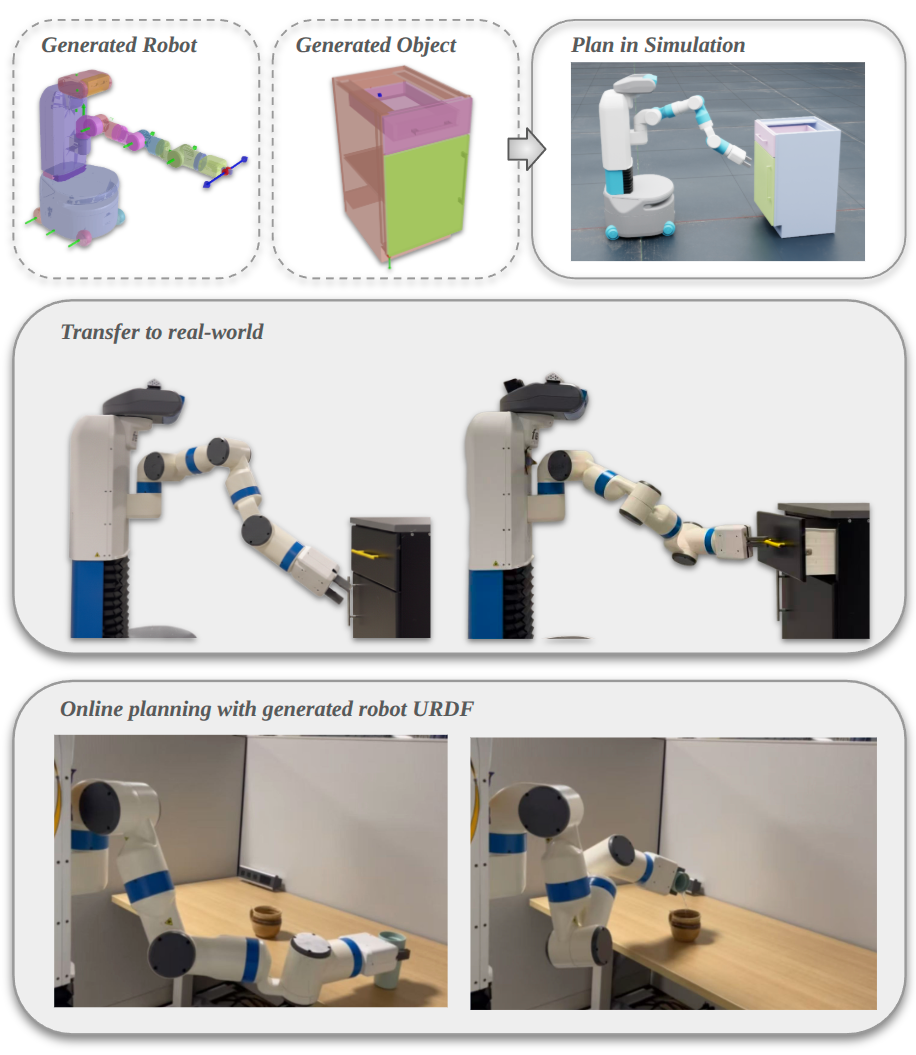}
\caption{\textbf{Real world experiment.} Kinematify generates URDFs for both the Fetch robot and the drawer, enabling a demonstration of the robot opening the drawer in Isaac Sim and transferring to real, with the same models usable for online planning with MoveIt \cite{chitta2012moveit}.}
\label{fig:real_world}
\end{figure}

\subsection{Real-world robot manipulation}

We export the recovered kinematics to URDF and deploy the models in simulation and on a real robot, shown in Fig.~\ref{fig:real_world}. From the segmented mesh, Kinematify generates a Fetch URDF and a simple cabinet URDF. For planning, we load both URDFs into a single MoveIt \cite{chitta2012moveit} planning scene and derive an SRDF group for the arm.

We use a constraint motion planner~\cite{hu2024motion, sucan2012ompl} as the backend. The task is executed in two stages: (1) reach-to-grasp and (2) constrained pull. In addition to this drawer-opening scenario, we also demonstrate an online planning task of pouring water from a cup into a container, using the same URDF pipeline and motion-planning setup. The same URDFs are used in Isaac Sim and on hardware. In both cases, the arm follows the planned trajectories without collision, showing that the recovered kinematics are physically consistent and directly usable for online planning in ROS and MoveIt \cite{chitta2012moveit}.



\section{Conclusion}
\label{sec:conclusion}

We presented \textbf{Kinematify}, an automated pipeline that synthesizes articulated object and robot descriptions from RGB images or text. Across everyday objects and multiple robot platforms, Kinematify improves joint estimation accuracy and kinematic tree fidelity over prior work.

Kinematify assumes accurate part segmentation and a reliable contact graph. In practice, spurious seams, missed contacts, and decorative geometry can mislead topology inference. Future work includes jointly refining segmentation and contact reliability during structure inference, and exploring a learning-based model trained on data generated by Kinematify that directly predicts kinematic topology and joint parameters from input, with physics-based constraints to ensure validity.

Overall, we view Kinematify as a step toward open-vocabulary synthesis of high-DoF articulated structures.

\section*{Acknowledgments}
\addcontentsline{toc}{section}{Acknowledgments}
This work was supported in part by the National Natural Science Foundation of China under Grant W2431046, National Key R\&D Program of China 2025YFA1309603, Central Guided Local Science and Technology Foundation of China YDZX20253100001001,and by MoE Key Lab of Intelligent Perceptionand Human-Machine Collaboration (ShanghaiTech University), the Shanghai Frontiers Science Center of Human-centered Artificial Intelligence.

\newpage
\bibliographystyle{ieeetr}
\bibliography{references}

\end{document}